\documentclass[11pt,a4paper]{article}
\usepackage[hyperref]{emnlp-ijcnlp-2019}
\usepackage{times}
\usepackage{latexsym}
\usepackage{graphicx}
\usepackage{subcaption}
\usepackage{todonotes}

\graphicspath{ {./images/} }

\usepackage{url}

\usepackage{color}

\newcommand{\gio}[1]{#1}
\newcommand{\del}[1]{{\color{orange}}}

\newcommand{\tanbih}{\textit{Tanbih}}

\aclfinalcopy 

\title{Tanbih: Get To Know What You Are Reading}

\author{Yifan Zhang$^1$, Giovanni Da San Martino$^1$, Alberto Barr\'on-Cede\~no$^2$, Salvatore Romeo$^1$, \\ \bf Jisun An$^1$, Haewoon Kwak$^1$, Todor Staykovski$^3$, Israa Jaradat$^4$, Georgi Karadzhov$^5$,\\ \bf Ramy Baly$^6$, Kareem Darwish$^1$, James Glass$^6$, Preslav Nakov$^1$\\
  $^1$Qatar Computing Research Institute, HBKU,
  $^2$Universit\`a di Bologna, Forl\`i, Italy\\
  $^3$Sofia University
  $^4$University of Texas at Arlington,
  $^5$SiteGround Hosting EOOD\\
  $^6$MIT Computer Science and Artificial Intelligence Laboratory\\
  \texttt{\{yzhang,gmartino,sromeo,jan,hkwak,kdarwish,pnakov\}@hbku.edu.qa}\\ \texttt{a.barron@unibo.it},
  \texttt{israa.jaradat@mavs.uta.edu}, \texttt{\{baly,glass\}@mit.edu}
}

\date{}

\begin{document}
\maketitle
\begin{abstract}
We introduce \tanbih, a news aggregator with intelligent analysis tools to help readers understanding what's behind a news story. 
Our system displays news grouped into events and generates media profiles that show the general factuality of reporting, the degree of propagandistic content, hyper-partisanship, leading political ideology, general frame of reporting, and stance with respect to various claims and topics of a news outlet. 
In addition, we automatically analyse each article to detect whether it is propagandistic and to determine its stance with respect to a number of controversial topics. \end{abstract}

\section{Introduction}

Nowadays, more and more readers consume news online. 
The reduced costs and, generally speaking, less strict regulations with respect to standard press, have led to a proliferation of the number of online sources. 
However, that does not necessarily entail that readers are exposed to a plurality of viewpoints. 
News consumed via social networks are known to reinforce the bias of the user~\cite{flaxman2016filter}. 
On the other hand, visiting multiple websites to gather a more comprehensive analysis of an event might be too time consuming for an average reader. 

News aggregators ---such as Flipboard\footnote{\url{https://flipboard.com}}, News Lens\footnote{\url{https://newslens.berkeley.edu}} and Google News\footnote{\url{https://news.google.com}.}---, gather news from different sources and, in the case of the latter two, cluster them into events. 
In addition, News Lens displays all articles about an event in a timeline and provides additional information, such as summary of the event and a description for each entity mentioned in an article. 

\noindent While these news aggregators help readers to get a more comprehensive coverage of an event, some of the sources might be unknown to the user, therefore he/she could naturally question the validity and trustworthiness of the information provided. 
Deep analysis of the content published by news outlets has been performed by expert journalists. For example, Media Bias/Fact Check\footnote{\url{http://mediabiasfactcheck.com}} provides reports on the bias and factuality of reporting of entire news outlets, whereas Snopes\footnote{\url{http://www.snopes.com/}} 
and FactCheck\footnote{\url{http://www.factcheck.org/}} are popular fact checking websites. 
All these manual efforts cannot cope with the rate at which news are produced. 
 
We propose \tanbih, a news platform that, in addition to displaying news grouped into events, provides additional information about the articles and their media source in order to develop the media literacy of users. 
Our system automatically generates media profiles with reports on the factuality, leading political ideology, hyper-partisanship, use of propaganda and bias of a news outlet. 
Furthermore, \tanbih~automatically categorizes articles in English and Arabic, flags potentially propagandistic ones, and examines framing bias. 

\section{System Architecture\label{sec:architecture}}

\begin{figure*}[t]
\centering
\includegraphics[width=0.68\textwidth]{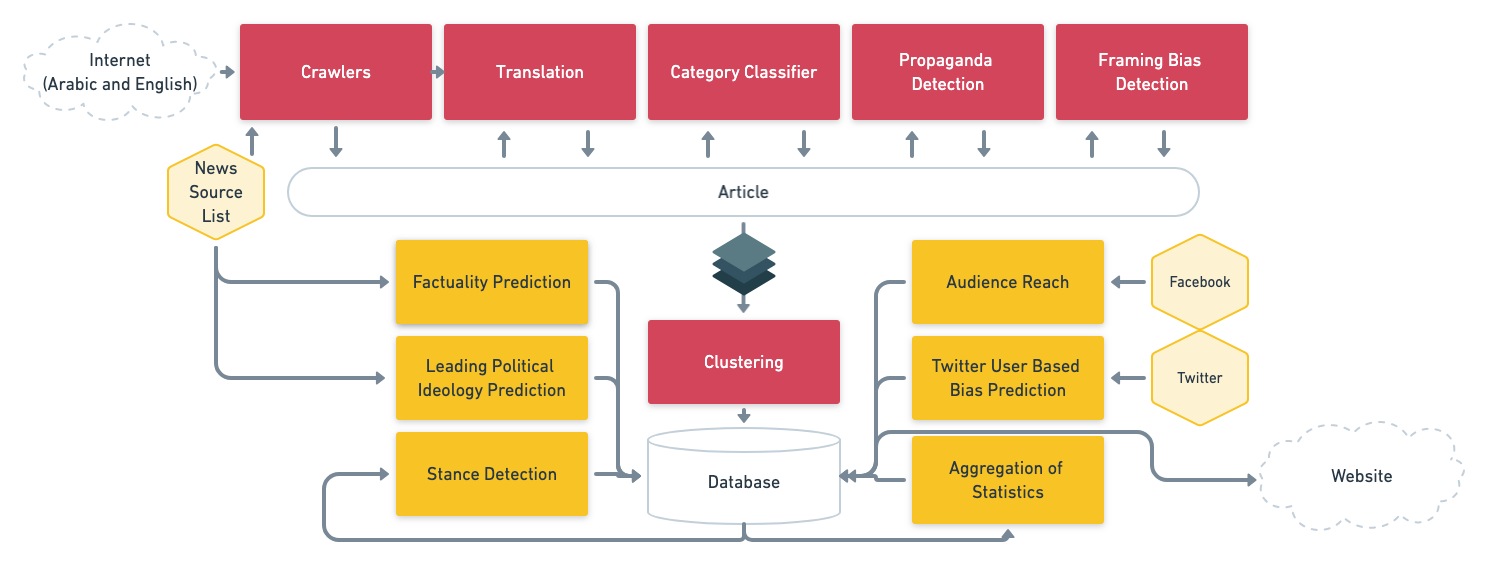}
\caption{Architecture of \tanbih. \label{fig:architecture}}
\end{figure*}

The architecture of \tanbih~is sketched in Figure~\ref{fig:architecture}. 
The system consists of three main components:
an online streaming processing pipeline for data collection and article level analysis, offline processing for event and media source level analysis and a website for delivering news to consumers. 
The online streaming processing pipeline continuously retrieves articles in English and Arabic. 
Translation, categorization, general frame of reporting classification and propaganda detection are performed for each article. 

\noindent Clustering is performed on the articles that are collected every 30 minutes. 
Offline processing includes factuality prediction, leading political ideology prediction, audience reach and twitter user based bias prediction on source level and stance detection, aggregation of statistics at article level, e.g. propaganda index (see Section~\ref{sec:propaganda}), for each medium. 
Offline processing does not have strict time requirements, therefore the choice of the models we develop will favour accuracy of the results over speed. 

In order to run everything in a streaming and scalable fashion, we use KAFKA%
\footnote{\url{https://kafka.apache.org}} 
as messaging queue and Kubernetes%
\footnote{\url{https://kubernetes.io}}
to manage scalability and fault-tolerant deployment.
In the following we describe each component of the system. 
\gio{We have open sourced the code for some of those, we will release the remaining ones upon acceptance of the corresponding research papers}. 


\subsection{Crawlers and Translation}
Our crawlers collect articles from an on-growing list of sources\footnote{\url{https://www.tanbih.org/about}}, which currently includes 155 RSS feeds, 82 twitter accounts and 2 websites. 
\gio{Once a link to an article is obtained from any of these sources, we rely on the Newspaper3k Python library to retrieve its content.\footnote{\url{https://newspaper.readthedocs.io}}}
After de-duplication, crawlers currently download 7k-10k articles every day. 
Currently we have more than 700k articles stored in our database. 
In order to display news both in English and in Arabic, we use QCRI Machine Translation~\cite{E17-3016} to translate English content into Arabic and vice versa. 
\gio{Since translation is performed offline, we select the most accurate system in~\citet{E17-3016}, i.e. the Neural-based one.}


\subsection{Section Categorization}
We build a model to classify an article into one of six news sections: Entertainment, Sports, Business, Technology, Politics, and Health. 
We build a corpus using the New York Times articles from the FakeNews dataset\footnote{https://github.com/several27/FakeNewsCorpus} published between Jan. 1st, 2000 and Dec. 31st, 2017.
We extract the news section information embedded in the article URL and in total we use 538k articles for training our models on TF-IDF representations of the contents. 
On a test set of 107k articles, the best-performing model is built based on Logistic Regression with F$_1$=$0.82, 0.58, 0.8,$ and $0.9$ for Sports, Business, Technology, and Politics, the sections used in our system, respectively. The baseline F$_1$ is 0.497. 

\subsection{Propaganda Detection} \label{sec:propaganda}

\del{News articles are might be purposefully biased to influence its readers and ultimately pursue a specific agenda. Such goal is reached by using propaganda techniques, which are designed to go unnoticed.} 
\del{News articles inevitably reflect, at some extent, the bias of the author and/or the publisher behind.  
Although that might happen unconsciously, in some cases an author is purposefully being biased to influence its readers and ultimately pursue a specific agenda. 
The latter case represents propaganda \cite{InstituteforPropagandaAnalysis1938}. 
Propaganda techniques are designed to go unnoticed, therefore spotting the use of propaganda in a text may not be an easy task for average readers. }
\del{We developed the propaganda detection component to flag articles that potentially could be propagandistic. }
\gio{We developed a propaganda detection component to flag articles that potentially could be propagandistic, i.e. purposefully biased to influence its readers and ultimately pursue a specific agenda. 
Given a corpus of news, binary labelled as propagandistic/non propagandistic~\cite{Barron:19}, we train a maximum entropy classifier trained on $51$k articles, represented with various style-related features, such as character $n$-grams and a number of vocabulary richness and readability measures, and obtain state-of-the-art F$_1$=$82.89$ on a separate test set of 10k articles. 
We refer to the score $p\in[0,1]$ of the classifier as \textit{propaganda index} and we define the following propaganda labels which we'll use to flag articles (see Figure~\ref{fig:interface}): very unlikely ($p<0.2$), unlikely ($0.2\leq p<0.4$), somehow ($0.4\leq p<0.6$), likely ($0.6\leq p<0.8$), and very likely ($p\geq 0.8$). } 

\del{The input to the propaganda identification module is one news article and the output is a \textit{propaganda index} $p\in[0,1]$. The index is translated in \tanbih~into one of the following labels:
very unlikely ($p<0.2$), unlikely ($0.2\leq p<0.4$), somehow ($0.4\leq p<0.6$), likely ($0.6\leq p<0.8$), and very likely ($p\geq 0.8$). 
The propaganda index is computed on the basis of a maximum entropy classifier trained on $51k$ articles represented with various style-related features, such as character $n$-grams and a number of vocabulary richness and readability measures. 
The model, tested on the binary task of detecting propagandistic versus non-propagandistic articles, has the state-of-the-art performance, estimated on a separate test set of 10k articles, of F$_1$=$82.89$. 
Refer to~\cite{Barron:19} for details. }


\subsection{Framing Bias Detection}

Framing is a central concept in political communication, which intentionally emphasizes (or ignores) certain dimensions of an issue~\cite{entman1993framing}. 
In \tanbih, we infer frames of news articles to make it transparent.
We use the Media Frames Corpus (MFC)~\cite{card2015media} for training our model to detect topic-agnostic media frames. 
%

We fine-tuned the BERT-based model  
with our training data using a small learning rate, 0.0002, a maximum sequence number of 128, and a batch size of 32. 
The performance of the model, trained on 11k articles in MFC, is an accuracy of 66.7\% on a test set of 1,138 articles, which is better than the reported state-of-the-art (58.4\%) from the subset of MFC~\cite{ji2017tree}. 

\subsection{Factuality of Reporting and Leading Political Ideology of a Source}
\gio{Verifying the reliability of the source is one of the basic tools used by investigative journalists to verify information reliability. 
To tacke this issue, we incorporated findings from our recent research on classifying the political bias and factuality of reporting of a news media~\cite{baly2018predicting} into \tanbih.
}
\del{While some of the analysis performed in \tanbih~is at the article level, we augment that information with insights on the factuality of reporting and the political ideology of the publishing source.
The hypothesis is that, if a website is known to have published non-factual information in the past, it is likely to do so in the future.
Furthermore, verifying the reliability of the source is one of the basic tools used by investigative journalists to verify information reliability.
To do so, we incorporated findings from our recent research on classifying the political bias and factuality of reporting of a news media~\cite{baly2018predicting} into \tanbih. }
In order to predict the factuality and the bias for a given news medium, we considered: 
a representation for a \textit{typical} article of a medium by averaging linguistic and semantic features of all articles of the medium; features extracted from the Wikipedia page of the source and from the metadata of the Twitter account, the structure of the medium's URL to identify malicious patterns~\cite{ma2009identifying} and web traffic through the {\it Alexa Rank}\footnote{\url{http://www.alexa.com/}}. 
\del{    {\bf Wikipedia page of the source.} We extracted a binary feature that indicates whether or not a Wikipedia page exists for the corresponding medium.
    Also, a semantic representation of the page was obtained by averaging the embeddings of the words it contains. Such representations should provide information about media's political ideology or alignment with certain political movements. 
    {\bf Twitter account.} We analyzed whether a news medium is active on Twitter or not, what is written in their description, and other relevant account metadata such as location, date of creation, etc.
    {\bf URL structure.} We analyzed the structure of the medium's URL to identify patterns of malicious URLs~\cite{ma2009identifying}.
    {\bf Web Traffic.} We used {\it Alexa Rank}\footnote{\url{http://www.alexa.com/}}, which helps detecting malicious websites that usually disappear after serving their purpose.}

In order to collect gold labels for training our supervised models, we used the data from the Media Bias/Fact Check (MBFC) website,\footnote{\url{https://mediabiasfactcheck.com}} which contains reliable annotations of factuality, bias and other aspects for over 2,000 news media. 
\del{We crawled articles, retrieved Twitter profiles and Wikipedia pages for each medium, in order to extract the above-mentioned features. 
These features are then used to train a separate Support Vector Machine (SVM) classifier to predict factuality or bias. }
\gio{We train a Support Vector Machine (SVM) classifier to predict factuality or bias using the representations above. }
Factuality of reporting was modeled at a 3-point scale ({\it low}, {\it mixed} and {\it high}), and the model achieved a 65\% accuracy.
On the other hand, political ideology was modeled on a {\it left}-to-{\it right} scale, and the model achieved a 69\% accuracy. 
\del{We are currently investigating multi-task models that predict both tasks jointly~\cite{baly-multitask-2019}.} 

\subsection{Stance Detection}
Stance detection aims to identify the relative perspective of a piece of text with respect to a claim, typically modeled using labels such as \textit{agree}, \textit{disagree}, \textit{discuss}, and \textit{unrelated}.
An interesting application of stance detection is medium profiling with respect to controversial topics. In this setting, given a particular medium, the stance for each article is computed with respect to a set of predefined claims. 
The stance of a medium is then obtained by aggregating the stance at article level. In \tanbih~the stance is used to profile media sources. 

We implemented our stance detection by fine-tuning the BERT classifier on the FNC-1 dataset from the \textit{Fake News Challenge}\footnote{\url{http://www.fakenewschallenge.org/}}. \gio{Our model outperforms the best submitted system}\del{and outperformed by a large margin the best known system}~\cite{Hanselowski:2018}. 
In particular, our system obtained F$_{1_{macro}}=75.30$ and F$_1=69.61, 49.76, 83.01,$ and $98.81$ for \textit{agree}, \textit{disagree}, \textit{discuss}, and \textit{unrelated} classes, respectively. 


\subsection{Audience Reach}

User interactions on Facebook enables the platform to generate comprehensive user profiles such as gender, age, income bracket, and political preferences. 
After marketers have determined a set of criteria for their target audience, Facebook can then provide them with an estimate of the size of this audience on its platform 
To illustrate, there are an estimated 160K Facebook users that are 20-year-old, very liberal females with an interest in The New York Times. 
\del{These target audience estimates, which constitute a source of `digital trace' data, are used across a variety of fields to measure online populations~\cite{ribeiro2018media}. }
In our system, we exploit the demographic composition, the political leaning in particular, of Facebook users who follow news media as a means to improve media bias prediction. 

To get the audience of each news medium, we use Facebook's Marketing API to identify the medium's ``Interest ID.'' Using this ID, we then extract the demographic data of the medium's audience with a focus on audience members who reside in the US and their political leanings (ideologies), which we categorize according to five classifications: ({\it Very Conservative, Conservative, Moderate, Liberal, and Very Liberal})\footnote{Political leaning information is only available for US-based Facebook users}.

\subsection{Twitter User-Based Bias Classification \label{sec:twitter}}
Controversial social and political issues may spur social media users to express their opinion through sharing supporting newspaper articles.  Our intuition is that the bias (or ideological leaning) of news sources can be inferred based on the bias of users.  For example, if articles from a news source are strictly shared by left or right leaning users, then the source is likely far-left or far-right leaning respectively.  Similarly, if it is being cited by both groups, then it is likely closer to the center. We used an unsupervised user-based stance detection method on different controversial topics to find core groups of right and left-leaning users \cite{darwish2019unsupervised}.  Given that the stance detection produces clusters with nearly perfect purity ($>$ 97\% purity), we used the identified core users to train a deep learning-based classifier, fastText \cite{joulin2016bag}, using the accounts that they retweeted as features to further tag more users.  Next, we computed the so-called valence score for each news source for each topic.  The valence scores ranges between -1 and 1, with higher absolute values indicating being cited with greater proportion by one group as opposed to the other. The score is calculated as follows \cite{conover2011political}: $V(u) = 2 \frac{\frac{tf(u, C_0)}{total(C_0)}}{\frac{tf(u, C_0)}{total(C_0)} + \frac{tf(u, C_1)}{total(C_1)}} - 1$, where $tf(u, C_0)$ is the number of times (term frequency) item $u$ is cited by group $C_0$, and $total(C_0)$ is the sum of the term frequencies of all items cited by $C_0$. $tf(u, C_1)$ and $total(C_1)$ are defined in a similar fashion.  We subdivided the range between -1 and 1 into 5 equal size ranges and assigned the labels \textit{far-left}, \textit{left}, \textit{center}, \textit{right}, and \textit{far-right} to the ranges.  

\subsection{Event Identification / Clustering}
The clustering module aggregates news articles into stories. The pipeline is divided in two stages: (i)  local topics identification and (ii) long-term topics matching for story generation.

For step (i), We represent each article as a $tf--idf$ vector, built from the title and the body concatenated. The pre-processing consists of casefolding, lemmatization, punctuation removal, and stopwording.
In order to obtain the preliminary clusters, in stage (i) we compute the cosine similarity between all article pairs in a predefined time window.
We set $n=6$ as the number of days withing a window with an overlap of 3 days.

The resulting matrix of similarities for each window is then used to build a graph $G = (V,E)$ where $V$ is the set of vertices ---the news articles--- and $E$ is the set of edges. An edge between two articles $\{d_i,d_j\}\in V$ is drawn only if $sim(d_i,d_j)\geq T_1$, with $T_1=0.31$. We select all parameters empirically on the training part of the corpus from~\cite{Miranda:18}.

The sequence of overlapping local graphs is merged in order of their creation, thus generating stories from topics.
After merging, a community detection algorithm is used in order to find the correct assignment of the nodes into clusters. We use one of the fastest modularity-based algorithms: the Louvain method~\cite{Blondel:08}. 

For step (ii), the topics created from the preceding stage are merged if the cosine similarity $sim(t_i,t_j)\geq T_2$, where $t_i$ ($t_j$) is the mean of all vectors belonging to topic $i$ ($j$), with $T_2 = 0.8$.

The model has state-of-the-art performance, on the test partition of the corpus from~\citet{Miranda:18}: \mbox{$F_1 = 98.11$} and F$_{1_{BCubed}} = 94.41$ (F$_{1_{BCubed}}$ is an evaluation measure specifically designed to evaluate clustering algorithms~\cite{Amigo:09}). 
As a comparison, the best model in~\cite{Miranda:18} has \mbox{$F_1 = 94.1$} (see ~\citet{Staykovski:19} for further details). 

\section{Interface}

\begin{figure}
\centering
\includegraphics[width=0.5\textwidth]{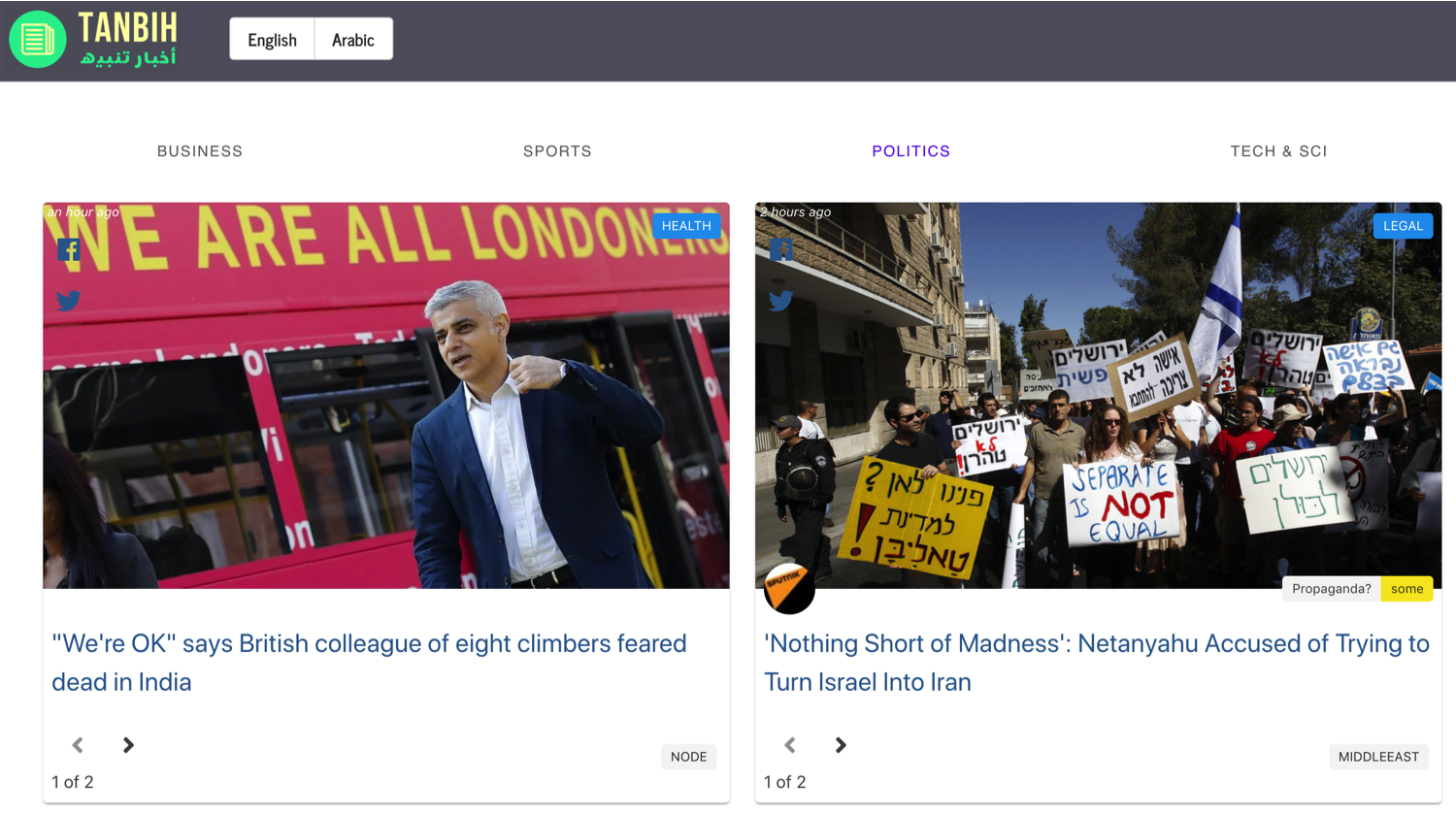}
\caption{\tanbih~website home page.}\label{fig:interface}
\end{figure}

\begin{figure*}
\centering
\begin{subfigure}{.32\textwidth}
    \centering
    \includegraphics[width=1\textwidth]{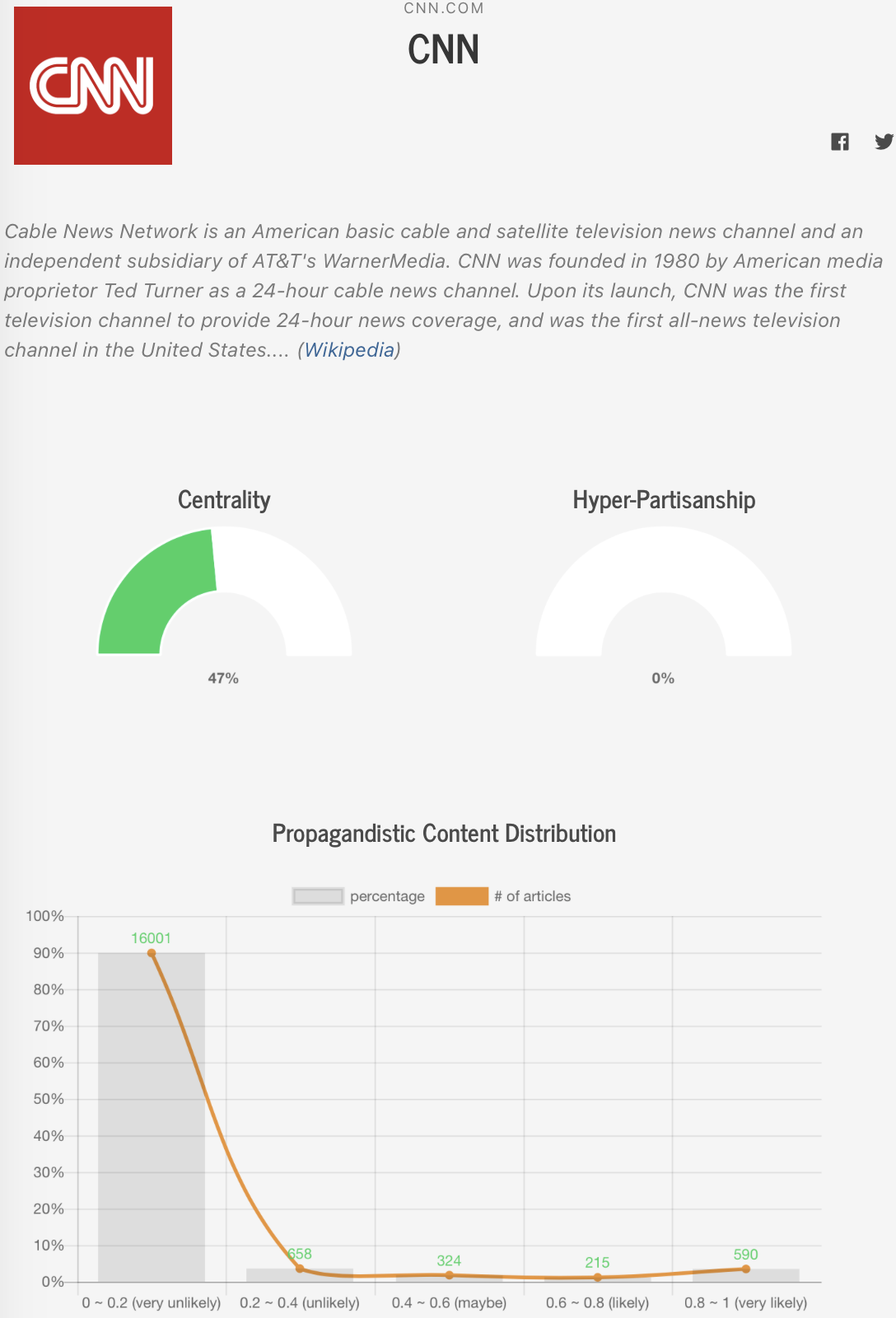}
    \caption{\label{fig:profilea}}
\end{subfigure}%
\begin{subfigure}{.32\textwidth}
    \centering
    \includegraphics[width=1\textwidth]{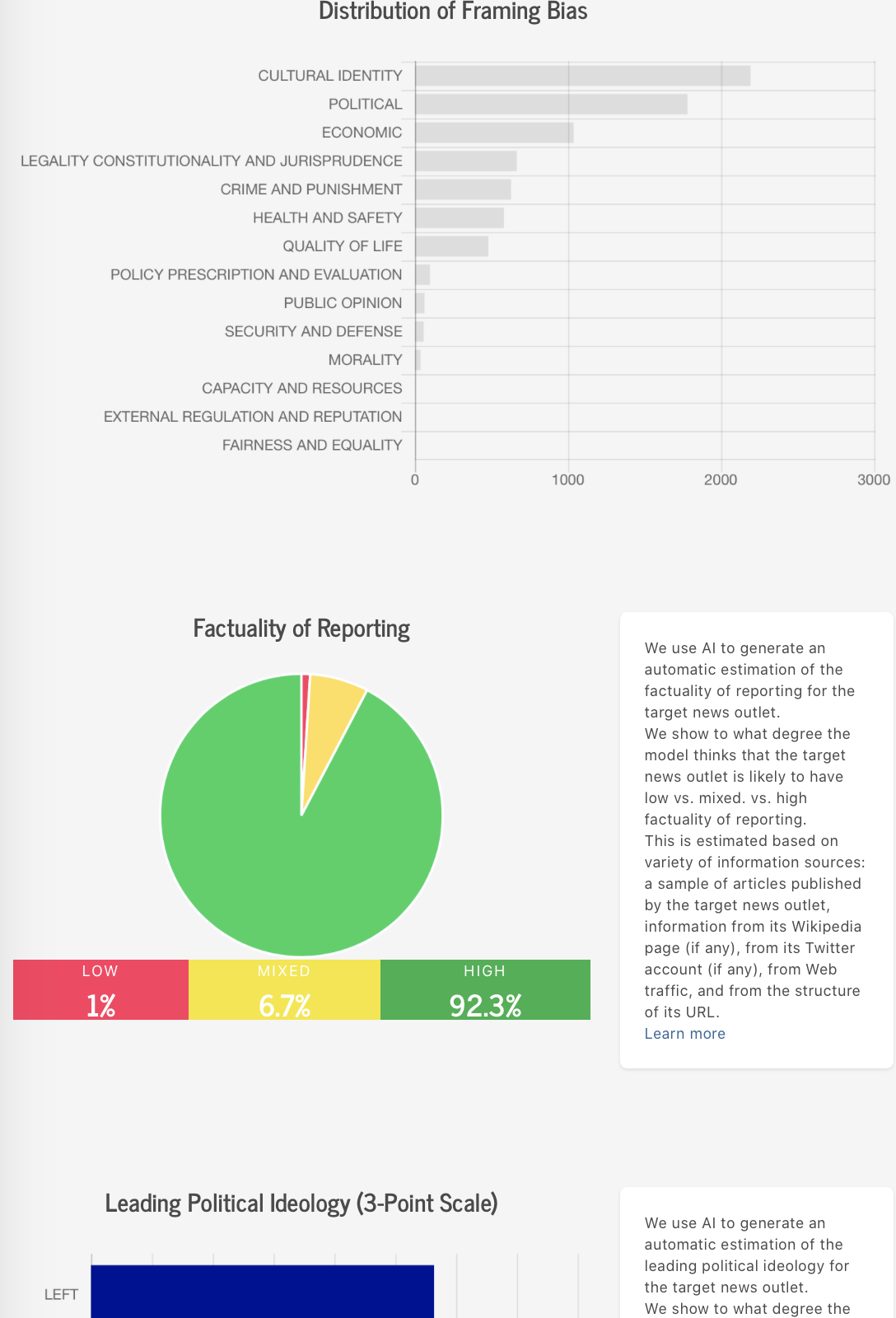}
    \caption{\label{fig:profileb}}
\end{subfigure}
\begin{subfigure}{.33\textwidth}
    \centering
    \includegraphics[width=1\textwidth]{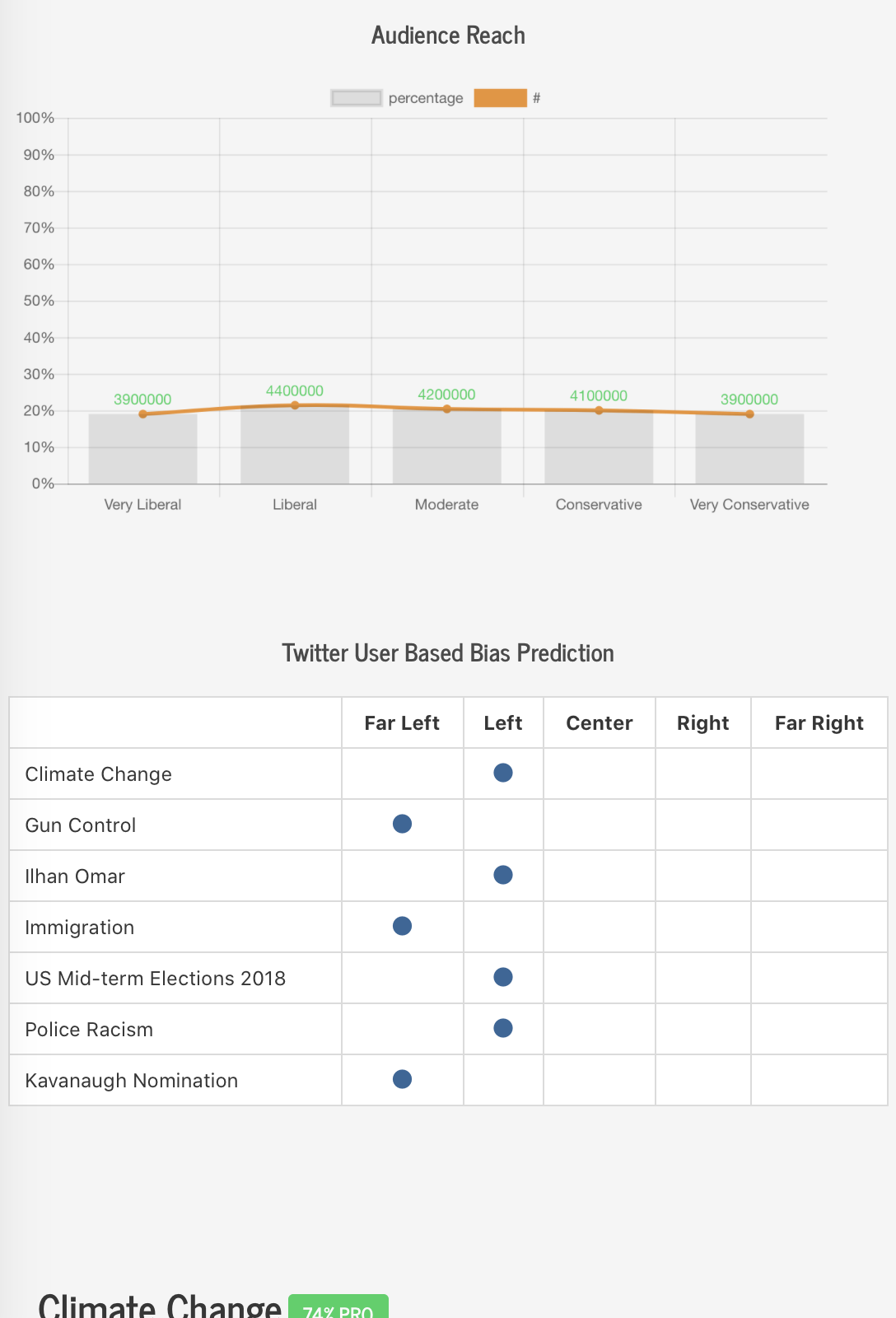}
    \caption{\label{fig:profilec}}
\end{subfigure}
\caption{A partial screenshot of the media profile page for CNN in \tanbih.}\label{fig:profile}
\end{figure*}

The home page of \tanbih\footnote{\url{http://www.tanbih.org}} displays news articles grouped into stories, i.e.,~clusters of articles (see the screenshot in Figure~\ref{fig:interface}).
Each story is displayed as a card. Users can go back and forth between the articles of an event by clicking on the left/right arrows below the title of the article. 
The propaganda label is displayed if the article is propagandistic. 

In Figure~\ref{fig:interface} the article from the Sputnik is flagged as likely to be propagandistic by our system. 
The source of each article is displayed with the logo or the avatar of the respective news organization, and it links to a profile page of this organization (see Figure~\ref{fig:profile}). 
On the top-left of the home page, \tanbih~ provides language selection buttons, currently English and Arabic only, to switch the language the news are display in. 
A search box in the top-right corner is also provided allowing the user to find the profile page of a particular news medium of interest. 

\begin{figure}[t]
\centering
\begin{subfigure}{.5\textwidth}
    \centering
    \includegraphics[width=1\textwidth]{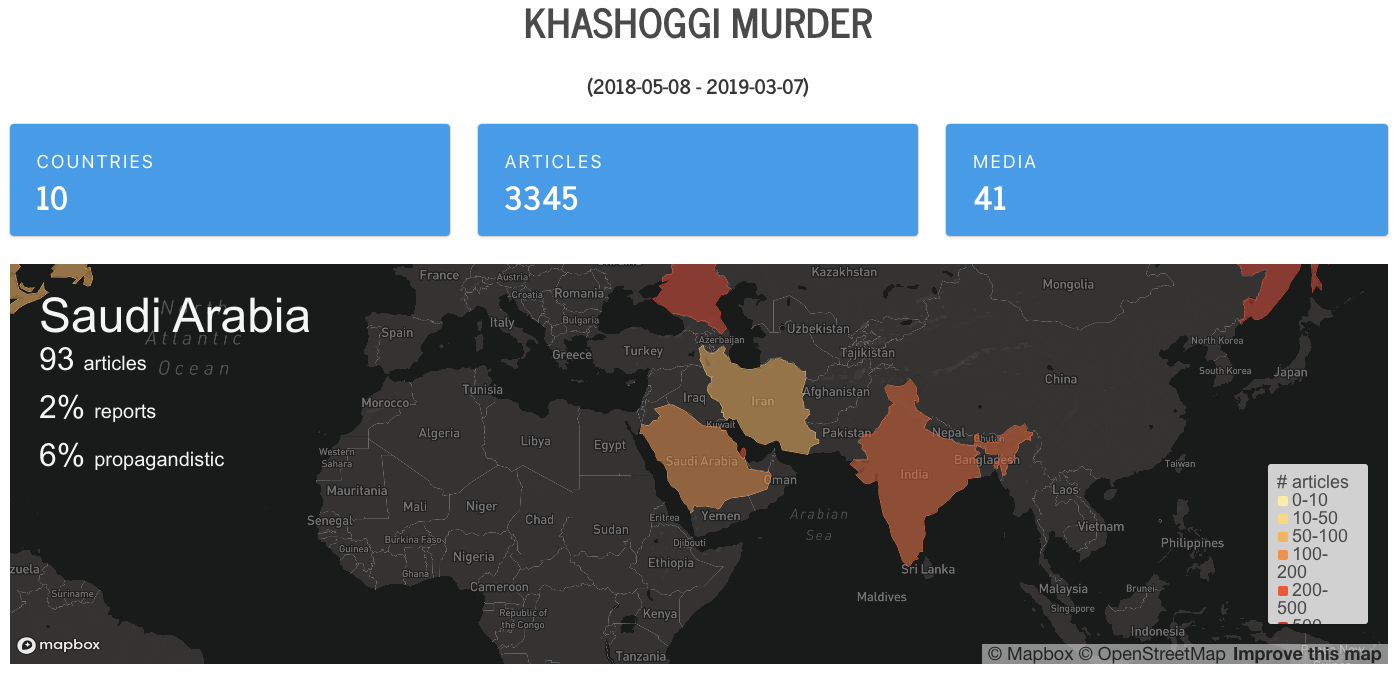}
\end{subfigure}
\begin{subfigure}{.5\textwidth}
    \centering
    \includegraphics[width=1\textwidth]{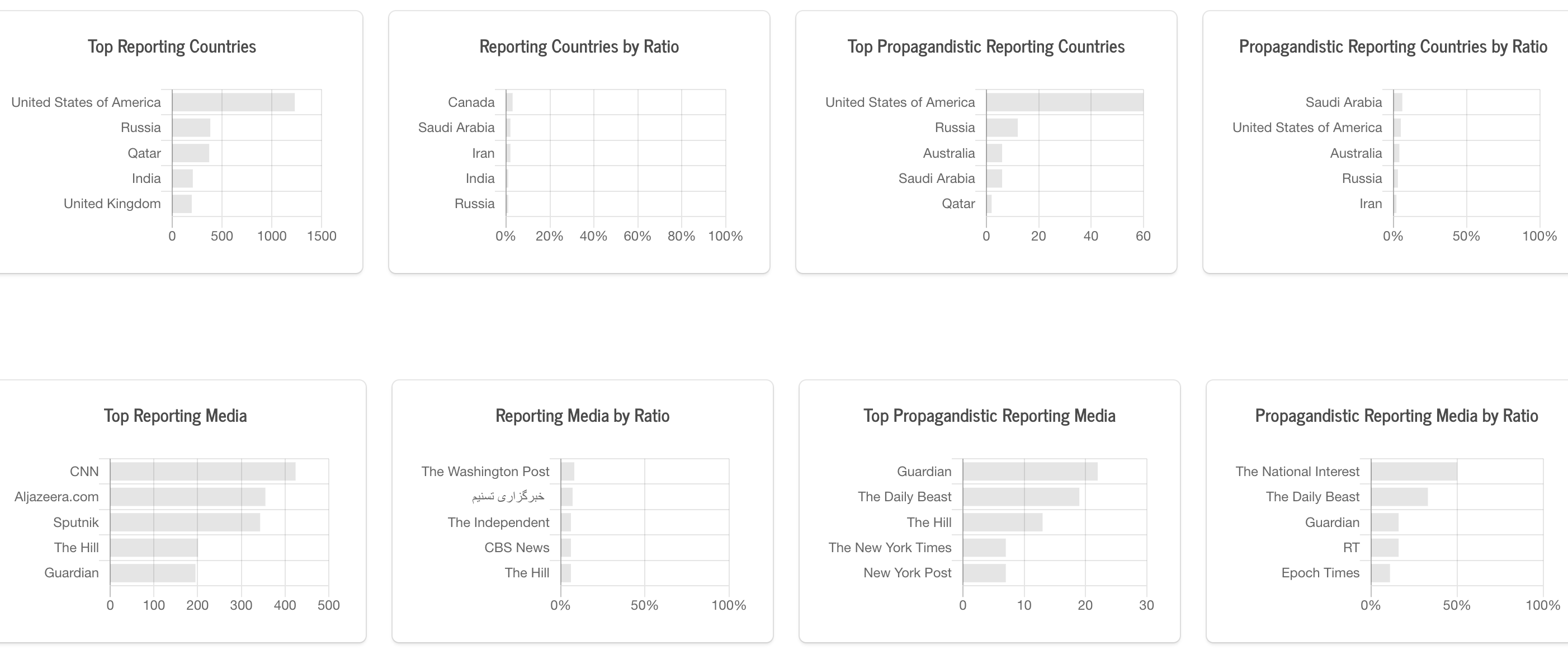}
\end{subfigure}
\caption{A partial screenshot of the topic page for Khashoggi Murder in \tanbih.}\label{fig:topic}
\end{figure}

On the media profile page (Figure~\ref{fig:profilea}), a short extract from the Wikipedia page of the medium is displayed on top, with recently-published articles on the right-hand side. 
The profile page includes a number of statistics automatically derived from the models in Section~\ref{sec:architecture}. We use as an example Figure~\ref{fig:profile} which shows screenshots of the profile of CNN\footnote{CNN full profile page is available at \url{https://www.tanbih.org/media/1}}. The first two charts in Figure~\ref{fig:profilea} show centrality and hyper-partisanship (in the example, CNN is reported as fairly central and low hyper-partisan) and the distribution of propagandistic articles (CNN publishes mostly non-propagandistic articles). 
Figure~\ref{fig:profileb} shows the overall framing bias distribution for the medium (CNN focuses mostly on cultural identity and politics), factuality of reporting (CNN is mostly factual). The profile also shows the leading political ideology distribution of the medium.  
Figure~\ref{fig:profilec} shows audience reach of the medium and the bias classification according to users' retweets (see Section \ref{sec:twitter}): CNN is popular among readers with any political view, although it tends to have a left-leaning ideology on the topics listed. 
The profile also features reports on the stance of CNN on a number of topics. 

Using the topic search box on the \tanbih~ home page, user can find the dedicated page of a topic, for example Brexit or the Khashoggi's murder. The top of the Khashoggi's murder event page is shown in Figure~\ref{fig:topic}. 

\noindent Recent stories in this topic will be listed on the top of the page, followed by statistics such as number of countries, number of articles and number of media. 
A map showing how much reporting on this event by each country allows users to have an overview of how important this topic is for these countries. 
The page also has two sets of charts showing \textit{i)} the top countries in terms of coverage of the event, both by absolute numbers and by ratio with respect to the total number of articles published; \textit{ii)} the media sources that have most propagandistic content on the topic, again both in absolute terms and by ratio with respect to the total number of articles published by the medium on the topic. 
The profile page also features plots equivalent to the ones in Figure~\ref{fig:profileb}, showing the distribution of propagandistic articles and the framing bias on a topic. 

\section{Conclusions and Future Work}

We have introduced \tanbih, our news aggregator which automatically computes media level and article level analyses to help the user in better understanding what they are reading. 
\tanbih~features factuality prediction, propaganda detection, stance detection, translation, leading political ideology analysis, media framing bias detection, and event clustering. 
The architecture of \tanbih~is flexible, fault-tolerant and it is able to scale to handle thousands of sources. 

As future work, we plan to expand the system to include many more sources, especially from non-English speaking regions and to add interactive components, for example letting users ask questions about a topic. 

\bibliography{main}
\bibliographystyle{acl_natbib}

\end{document}